\documentclass{article}

\usepackage[final,nonatbib]{neurips_2023}

\usepackage[utf8]{inputenc} 
\usepackage[T1]{fontenc}    
\usepackage{hyperref}       
\usepackage{url}            
\usepackage{booktabs}       
\usepackage{amsfonts}       
\usepackage{nicefrac}       
\usepackage{microtype}      
\usepackage{xcolor}         
\usepackage{pifont}
\usepackage{appendix}
\usepackage{caption}
\usepackage{listings}
\usepackage{soul}
\usepackage{authblk}
\usepackage{graphicx}
\graphicspath{ {./figs/} }

\usepackage[style=numeric, backend=biber, sorting=none]{biblatex}
\title{Large Language Models with Retrieval-Augmented Generation for Zero-Shot Disease Phenotyping}

\author[1]{\textbf{Will E. Thompson, MS}}
\author[1]{\textbf{David M. Vidmar, PhD}}
\author[1]{\textbf{Jessica K. De Freitas, PhD}}
\author[1]{\authorcr \textbf{John M. Pfeifer, MD, MPH}}
\author[1]{\textbf{Brandon K. Fornwalt, MD, PhD}}
\author[1]{\textbf{Ruijun Chen, MD, MS}}
\author[1]{\authorcr \textbf{Gabriel Altay, PhD}}
\author[1]{\textbf{Kabir Manghnani, BS}}
\author[2]{\textbf{Andrew C. Nelsen, PharmD}}
\author[2]{\authorcr \textbf{Kellie Morland, PharmD}}
\author[1]{\textbf{Martin C. Stumpe, PhD}}
\author[1]{\textbf{Riccardo Miotto, PhD}}
\affil[1]{Tempus Labs, Inc.\\
600 West Chicago Avenue\\
Suite 510\\ 
Chicago, IL 60654 USA}
\affil[2]{United Therapeutics Corporation\\
1000 Spring Street\\
Silver Spring, Maryland 20910 USA}

\lstnewenvironment{regex}[1][]{
    \lstset{
        language=TeX,
        basicstyle=\ttfamily,
        escapeinside={(*@}{@*)}, 
        #1 
    }
}{}

\begin{document}

\maketitle

\begin{abstract}
Identifying disease phenotypes from electronic health records (EHRs) is critical for numerous secondary uses. Manually encoding physician knowledge into rules is particularly challenging for rare diseases due to inadequate EHR coding, necessitating review of clinical notes. Large language models (LLMs) offer promise in text understanding but may not efficiently handle real-world clinical documentation. We propose a zero-shot LLM-based method enriched by retrieval-augmented generation and MapReduce, which pre-identifies disease-related text snippets to be used in parallel as queries for the LLM to establish diagnosis. We show that this method as applied to pulmonary hypertension (PH), a rare disease characterized by elevated arterial pressures in the lungs, significantly outperforms physician logic rules ($F_1$ score of 0.62 vs. 0.75). This method has the potential to enhance rare disease cohort identification, expanding the scope of robust clinical research and care gap identification.\end{abstract}

\section{Introduction}

Disease phenotyping within electronic health records (EHRs) involves identifying ground truth diagnoses in a patient's clinical history. These phenotypes play a crucial role in several essential functions, such as selecting patient groups for observational studies or interventional quality initiatives to close gaps in care, defining inclusion and exclusion criteria, and providing labels for subsequent modeling tasks. Relevant information for disease diagnosis may be scattered across EHRs different data sources, including free-text notes, International Classification of Diseases 9th and 10th revision (ICD-9/10 CM) codes, medications, or laboratory values from medical procedures and tests. 

The ideal process involves subject matter experts (SMEs) reviewing patient files for disease diagnosis. Yet chart reviews are time-consuming, taking an average of 30 minutes per file \cite{mckenzie_semiautomated_2021}. To address this, SMEs often create rules-based algorithms, combining ICD codes, laboratory values, medications, and procedures, to identify diseases. However, challenges arise, including coding errors, reporting biases, and data sparsity, requiring iterative refinement through a human-in-the-loop process. Scalability is hindered, especially when features from one EHR system do not generalize to others \cite{ostropolets_reproducible_2023}. Mapping rare diseases to common ontologies can be especially problematic due to expert disagreements \cite{fung_coverage_2014, mortensen_empirically_2014}.

Machine learning phenotyping approaches, both supervised and unsupervised, have shown varying degrees of promise \cite{ho_limestone_2014, pivovarov_learning_2015, halpern_electronic_2016, glicksberg_automated_2018, ahuja_surelda_2020, de_freitas_phe2vec_2021}. Supervised learning approaches often require high-quality labels and are therefore constrained by a labeling bottleneck. While unsupervised learning approaches circumvent this problem, they often are difficult to tailor to a particular disease definition or achieve certain acceptance criteria. The majority of work also mostly focuses on structured EHRs; incorporating the notes in a phenotype definition holds the possibility of developing a better model \cite{vidmar_abstract_2022, moldwin_empirical_2021}. 

To overcome these limitations within practical computational constraints, this paper introduces a retrieval-augmented generation (RAG) approach to zero-shot phenotyping using large language models (LLMs). To our knowledge, this is the first study to explore the application of a RAG approach to process entire patient records with LLMs, as opposed to focusing solely on a specific type of clinical note. An essential aspect of this retrieval step is its ability to analyze all clinical mentions throughout a patient's entire record without the need for predefined sections of interest. Given the substantial volume of potentially relevant information retrieved, and the aim to assess the language model's reasoning abilities rather than relying on an intricate retrieval mechanism, this paper also investigates a MapReduce paradigm for parallel snippet evaluation and resolution of potentially conflicting information during the output aggregation stage. To assess the performance of this approach, we enlist a physician SME to assist in developing a competing rules-based model, the commonly used approach in healthcare practice and industry. Both models are then evaluated using an unseen test dataset, with our chosen disease phenotype being pulmonary hypertension (PH).

\section{Methodology}

\paragraph{LLM Phenotyping Pipeline}

\begin{figure}[ht]
    \centering
    \includegraphics[scale=0.40]{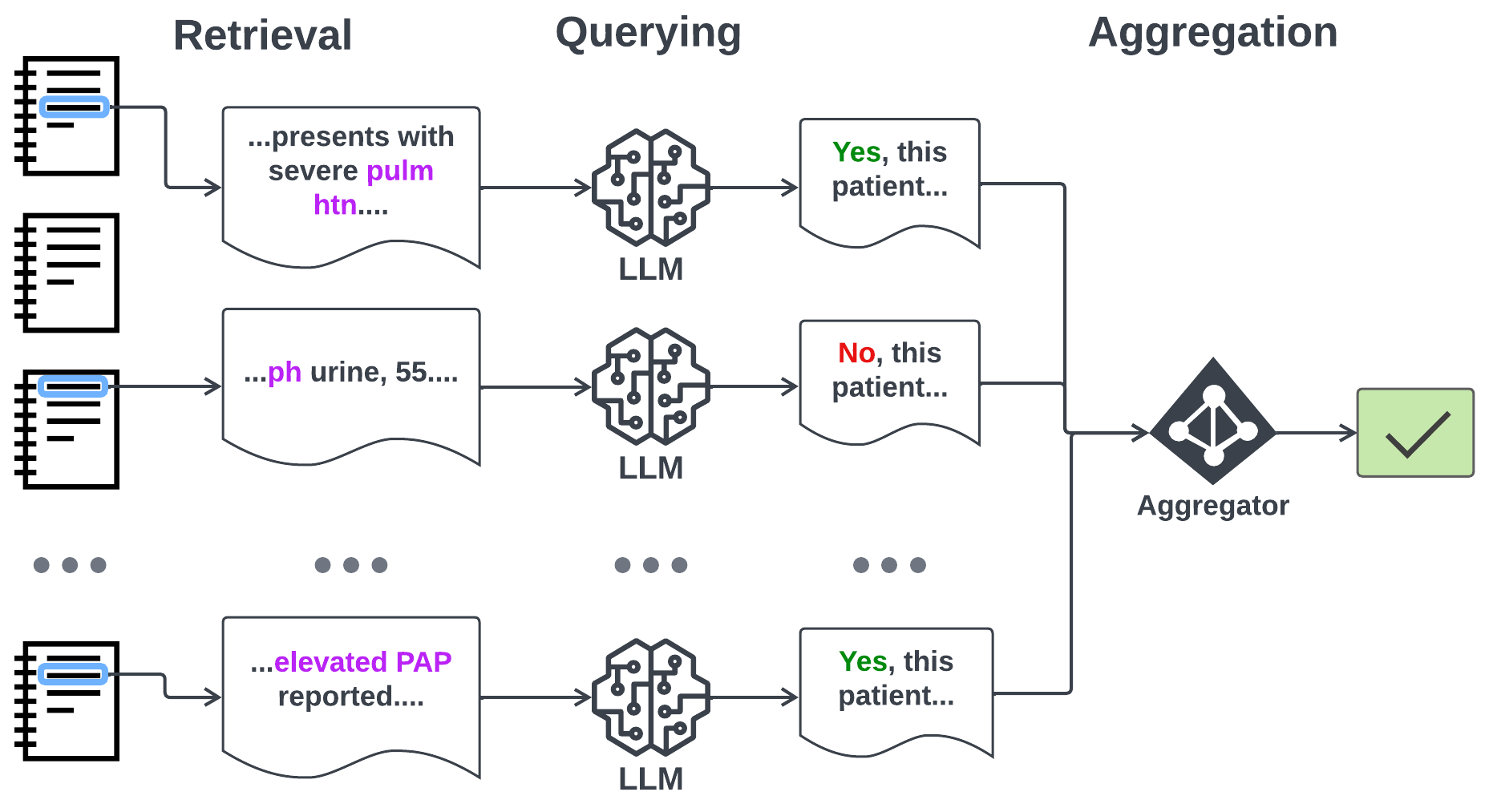}
    \caption{Overview of the LLM-based phenotype architecture.}
    \label{fig:pipeline}
\end{figure}

Figure \ref{fig:pipeline} shows the overall architecture of the LLM-based phenotyping model: a RAG approach followed by a MapReduce step for querying and aggregation.

\paragraph{Retrieval}

Considering the extensive volume of text contained within a real world data (RWD) warehouse of EHRs, it becomes impractical to process the entirety of a patient's clinical notes within the context window of an LLM. To address this challenge, we implemented a RAG approach based on regular expressions (Regex) to identify relevant snippets from the text. As shown in the literature, a RAG approach proves to be more efficient and effective than providing the LLM with larger context windows \cite{catav_less_2023}. We opted for Regex over more complex retrieval models, such as Term Frequency-Inverse Document Frequency, Cohere's re-rank \cite{reimers_say_2023}, or Instructor embeddings \cite{su_one_2023}, in order to avoid introducing bias through additional hyperparameter tuning and to narrow the focus of our work to assessing the LLM's capability in diagnosing diseases. Although building Regex queries requires domain knowledge which may not be easily extensible to all other disease contexts, we believe that this approach injects useful priors into the model pipeline. Further, some form of retrieval method is necessary as passing the entirety of a patients' clinical notes through an LLM is \textit{intractable} given the sheer size of clinical notes, even in the context of a MapReduce paradigm. We divided each clinical note into snippets of 2,048 “tokens” (i.e., the vocabulary of characters, words or subwords an LLM uses to partition free text) in size and then employed Regex to identify relevant snippets. Regex rules encompassed a broad spectrum of patterns that could potentially be associated with any mention of a specific medical condition, such as PH (Appendix \ref{sec:appendixA1}). Subsequently, all Regex-retrieved patient snippets are fed into the LLM to determine the diagnosis status.

\paragraph{Querying}

While the RAG approach reduces the amount of text processed by the LLM, RWD clinical notes often comprise many pages of text. Consequently, the Regex retriever is still likely to return a large number of snippets, which may exceed the LLM's context window. To address this, we employed a MapReduce approach where each snippet is concurrently presented as context to the LLM, along with a set of instructions to facilitate decision-making (i.e., the "map" phase). Generated outputs, one per snippet, are then aggregated to form the final decision (i.e., the "reduce" phase).  

\paragraph{Aggregation}

We evaluated two aggregation approaches for the “reduce” phase: (1) an LLM-based approach, which aggregates the output and reasoning from each individual snippet query into a larger prompt for a final decision by an LLM through prompting; and (2) a Max aggregation function, which checks if any of the individual snippet queries returned a positive diagnosis and, if so, assigns a positive label to the patient as a whole. In the LLM aggregation approach, we explored two variations: (1) apply the same prompt that was provided at  the snippet level to aggregate responses; and (2) apply a different prompt that asks the LLM if any of the responses indicated a positive diagnosis.  

\section{Evaluation Design}

\paragraph{Pulmonary Hypertension}

PH is a condition characterized by abnormally elevated pressure in the arteries of the lung and right side of the heart. While the prevalence of specific PH etiologies may vary across groups, it is broadly categorized as a rare disease with an estimated global prevalence rate of 1-3\% \cite{moreira_prevalence_2015}. Building a PH phenotype is complicated by the fact that the hemodynamic definition of PH has changed over time \cite{simonneau_haemodynamic_2019}. This means that there are some patients who would be currently identified as having PH under the new definition, would not have been diagnosed with PH during the time of their original treatment or workup. The ability to systematically identify PH patients who would otherwise not be identified could significantly impact patient outcomes. 

\paragraph{Dataset}

We utilized de-identified clinical notes from a medium-sized health system serving a population of approximately 2.2 million patients (as of December 2022). Given the expected low prevalence rate of PH within this population, we created an enriched cohort of patients displaying any clinical evidence of PH in either the structured data or clinical notes. From this cohort, we randomly selected 299 patients for a comprehensive chart review\footnote{We originally sampled 300 patients for curation; however, due to data integration issues, we dropped one patient during the curation process.}. Each patient underwent independent evaluation by two physicians, with any discrepancies resolved through joint discussion to reach a consensus. The cohort consisted primarily of a white population (97\%), comprised of 154 females and 145 males. The median age at the last encounter for each patient was 72 years (interquartile range: 60 to 84), and the median observation time, defined as the number of years between each patient's first and last encounter, was 17.5 (interquartile range: 8.75 to 22.92).
Subsequently, these labeled patients were divided into three groups: a training set comprising 50 patients (19 cases and 31 controls); a validation set also consisting of 50 patients (19 cases and 31 controls); and a test set with 199 patients (76 cases and 123 controls). 

\paragraph{Large Language Model}

We utilized ‘bison@001’, which is a version of PaLM-2 available through Google Cloud's Vertex-AI offering \cite{chowdhery_palm_2022}. As of the time of writing, this stands as the second-largest commercially available variant of the PaLM-2 model intended for consumer use. PaLM-2 builds upon the foundation of PaLM-1, incorporating a combination of various pre-training objectives to achieve state-of-the-art results on several benchmarks while maintaining smaller model sizes \cite{anil_palm_2023}.

\paragraph{Structured Phenotype Definition}

One of our internal physicians conducted a review of patients within the training dataset to establish a rules-based algorithm for diagnosing PH using EHRs. Following a thorough examination of the literature on PH phenotypes, the rules encompassed a blend of ICD-9/10 code frequencies, medication records, laboratory data, and other clinical features available in the patients' records. After a series of iterative reviews and adjustments to the model output, the physician ceased further model development when the incremental improvements began to diminish. The final phenotype is reported in Appendix \ref{sec:appendixA2} Table \ref{tab:structured_phenotype}.

\section{Results}

We conducted experiments involving several iterations of prompt design, snippet exclusions and aggregation methods. After evaluating the results using training and validation sets, we selected a subset of configurations for final assessment on the test set. The complete set of results on the validation set is presented in Appendix \ref{sec:appendixA3} Table \ref{tab:validation_results}. All performances are in terms of $F_1$ score.

\paragraph{Prompt Design Analysis}
 
We explored various zero-shot prompt designs to query the LLM for the diagnosis of PH. Additionally, we evaluated the value of Chain-of-Thought (CoT) reasoning by enhancing the prompt with the phrase "let's think step-by-step" \cite{wei_emergent_2022, kojima_large_2023}. False positives and false negatives were manually reviewed while iterating on prompt design. From this analysis, we concluded that our prompts needed to explicitly guide the model to consider possible cases of PH as negative diagnosis and history of PH as positive diagnosis. We found a fair amount of variability in performance across these different prompt designs without any prominent features defining those prompts that appeared to perform best. Therefore we chose the prompt design presented in Figure \ref{fig:prompt}, which is composed of CoT, multiple choice, history of PH and possible PH steering. See Appendix \ref{sec:appendixA3} Table \ref{tab:prompt_description} for the description of each prompt evaluated.
 
\begin{figure}[h]
    \centering
    \includegraphics[scale=0.40]{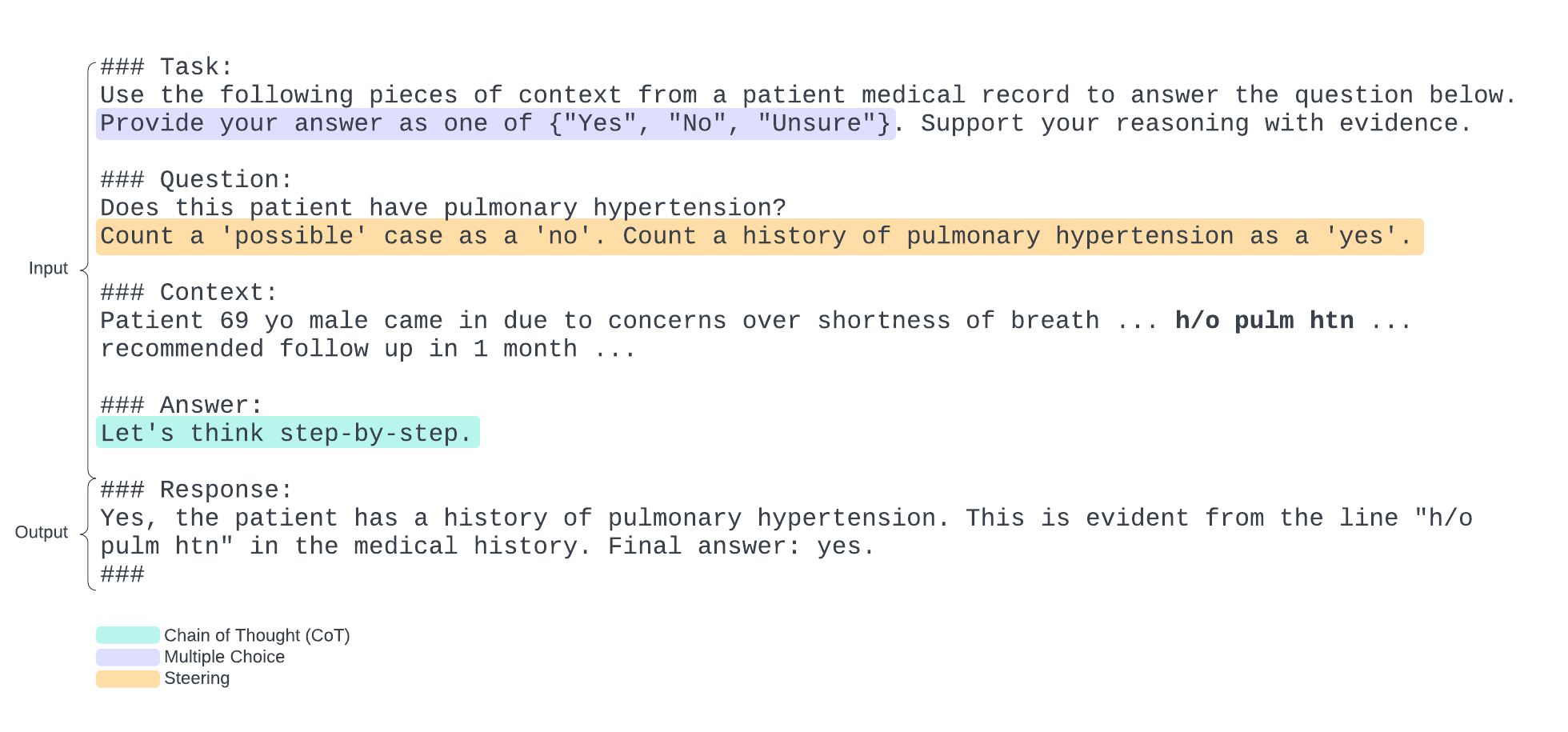}
    \caption{An illustrative example of zero-shot prompt.}
    \label{fig:prompt}
\end{figure}

\paragraph{Retrieval Exclusions}

While reviewing model errors on the validation set, we observed that a noteworthy percentage of false positives originated from echocardiogram (ECHO) or computerized tomography (CT) reports noting \textit{suspicion} of PH without any confirmatory diagnostic testing nor a clinical diagnosis by a provider. This was not surprising since it is known that both ECHO and CT have higher error rates when compared to the gold standard diagnostic i.e. right heart catheterization (RHC) \cite{amsallem_addressing_2016, rich_inaccuracy_2011, fisher_accuracy_2009, shen_ct-base_2014}. As these reports typically exhibit a consistent structure, we explored the possibility of excluding snippets extracted from the reports in two ways: (1) employing regular expressions to filter out snippets containing headers and common language found in these reports and (2) enhancing our prompts to instruct the LLM to disregard ECHO and CT reports. Results on the validation set indicated improvements in the performance of the model when excluding these snippets; therefore, we decided to include this filter in the final architecture. We also noticed that the structured phenotype definition did not include ECHO reports as it was built solely using structured EHRs and not clinical notes.

\paragraph{Aggregation Evaluation}

LLM-based aggregation methods, whether using the same or different prompts at the snippet level, resulted in average $F_1$ scores of 0.67 and 0.68, respectively, with significant variability in performance across choice of prompting design. In contrast, Max aggregation achieved an $F_1$ score of 0.73, also providing increased stability across different prompts.

\paragraph{Comparing LLM and Structured Phenotypes}

Table \ref{tab:results-tab} compares the performance of three variations of the LLM-based phenotyping architecture with the structured phenotype baseline developed by a physician on the test set. As demonstrated, LLM-based phenotypes generally show improvements between 18\% and 21\% over the structured phenotype. We did observe a notable drop in $F_1$ scores, ranging from 0.05 to 0.1, in the performance of LLM-based methods compared to the results obtained on the validation set (see Appendix \ref{sec:appendixA2} Table 2), which might be attributed to the larger evaluation cohort and potentially to some overfitting on the training set. Nevertheless, the LLM-based methods significantly outperformed the structured phenotype method, resulting in the identification of approximately twice as many patients with a confirmed diagnosis of PH. In a real-world application, these patients might otherwise remain unidentified. Furthermore, we observed that our retrieved documents spanned 29 distinct note types, highlighting the importance of retrieving across note types to accurately identify disease diagnoses (see Appendix \ref{sec:appendixA4} Table \ref{tab:note_types}).

\begin{table}[h]
  \centering
  \begin{tabular}{cccc}
    \toprule
    Model     & Aggregation  &  ECHO Exclusion & $F_1$ score  \\
    \midrule
    Structured & ---  &  ---             & 0.62     \\
    LLM        & Max  &  Regex& 0.73       \\
    LLM        & Max  &  Prompt Amended  & \textbf{0.75} \\
    LLM        & LLM  &  Prompt Amended  & 0.72  \\
    \bottomrule
  \end{tabular}
    \vspace{5mm}
    \caption{Test set performance of three variants of the LLM-based phenotype compared to structured phenotype.}
      \label{tab:results-tab}
\end{table}

\section{Conclusion}
This paper underscores the potential of employing an LLM-based architecture to identify diseases across clinical notes. Unlike existing literature, which often utilizes LLMs on specific types of notes, our method harnesses RAG and MapReduce to effectively analyze the complete patient documentation. Our experiments demonstrated the superiority of this method over SME rule-based models in diagnosing PH. Efficient LLM-based phenotype models offer scalability and improvement in identifying specific diseases in real-world EHRs, reducing the manual workload for SMEs and the need for ad-hoc machine learning models while enabling comprehensive patient record analysis. This advancement promises to enhance systems utilizing EHRs for purposes such as clinical decision support, care gap detection / population health management, clinical trial matching, and cohort generation. Future work will explore the integration of this method with structured data, the evaluation of other LLM engines, SME-enabled active learning loops, more advanced retrieval methods and additional clinical domains.

\pagebreak
\printbibliography

\pagebreak

\appendix

\section{Appendix A}

\subsection{Regular Expression (Regex) Retrieval}
\label{sec:appendixA1}
We present below the regular expression rules utilized to identify pertinent text snippets for diagnosing pulmonary hypertension (PH) within the retrieval-augmented generation (RAG) approach.
\label{regex}
\begin{regex}
    r'(?i)(?:\bPulm.{0,10}hypertension\b|\bPH\b|\bPulm.?HTN
    \b|\bp.?HTN\b|\bp.?AH\b|\barterial.hypertension\b|\bPHT\b|
    (?:\belevated\b|\bhigh\b).(?:\bPASP\b|\bpulm.{0,10}art.
    {0,5}sys|\bpulm.{0,10}art.{0,5}pr|\bPAP\b)|(?:\belevated\b|
    \bhigh\b).(?:\bRVSP\b|\bRVP\b|\br.{0,5}v.{0,15}sys.{0,7}
    pressure\b|\br.{0,5}v.{0,15}pressure\b)\b|\bflat.
    {0,7}septum\b|\bseptal.flat|(?:\benlarge.{0,15}|
    \bdilat.{0,15})\bpulm.{0,10}art|\bPH-ILD\b|\bPHILD\b|
    \bCTEPH\b|\bPH-COPD\b|\bPHCOPD\b)')
\end{regex}
These terms were curated by a physician SME using their domain knowledge.  Future work could explore automated, or semi-automated, data mining methods to curate or expand relevant terms.

\subsection{Pulmonary Hypertension Structured Phenotype}
\label{sec:appendixA2}
\begin{table}[ht]
    \centering
    \begin{tabular}{c|l|l}
    \toprule
        Code Vocabulary & Code & Name  \\
        \midrule
         & 416  & Chronic pulmonary heart disease \\
          & 416.0  & Primary pulmonary hypertension \\
        ICD-9 & 416.1 & Kyphoscoliotic heart disease \\
         & 416.2 & Chronic pulmonary embolism \\
         & 416.8 & Other chronic pulmonary heart diseases \\
          & 416.9 & Chronic pulmonary heart disease, unspecified \\
        \midrule
         & I27.21 & Secondary pulmonary arterial hypertension \\
         & I27.22 & Pulmonary hypertension due to left heart disease \\
          & I27.23 & Pulmonary hypertension due to lung diseases and hypoxia \\
        ICD-10 & I27.24 & Chronic thromboembolic pulmonary hypertension \\
          & I27.29 & Other secondary pulmonary hypertension \\
          & I27.8 & Other specified pulmonary heart diseases \\
          & I27.89 & Other specified pulmonary heart diseases \\
         & I27.9 & Pulmonary heart disease, unspecified \\
        \midrule
          & 1439816 & Ambrisentan \\
         & 1439816 & Bosentan \\
          & 8814 & Epoprostenol \\
          & 40138 & Iloprost \\
          & 1442132 & Macitentan \\
        RxNorm & 1439816 & Riociguat \\
          & 1729002 & Selexipag \\
          & 136411 & Sildenafil \\
          & 358263 & Tadalafil \\
          & 343048 & Treprostinil \\
          \midrule
    \end{tabular}
    \vspace{2mm}
    \caption{Diagnostic and medication codes that make up the structured phenotype for PH.}
    \label{tab:structured_phenotype}
\end{table}

\subsection{Validation Experimental Results}
\label{sec:appendixA3}
This section presents the phenotyping results of the LLM-based architecture for the various configurations assessed in the validation dataset (Table \ref{tab:validation_results}) and enumerates the different types of prompt strategies that were considered (Table \ref{tab:prompt_description}).

\begin{table}[ht]
    \tiny   
    \centering
\begin{tabular}{c ccc ccc ccc}
\toprule
 & \multicolumn{3}{c}{LLM - Same Prompt} & \multicolumn{3}{c}{LLM - Different Prompt} & \multicolumn{3}{c}{Max Aggregation} \\
\cmidrule(lr){2-4} \cmidrule(lr){5-7} \cmidrule(lr){8-10} 
Prompt     & No excl.   & Regex excl. & Prompt excl. & No excl. &  Regex excl. & Prompt excl. & No excl. & Regex excl.  & Prompt excl.  \\
\midrule
A       & 0.67 & 0.69 & \textbf{\textcolor{red}{0.82}} & 0.71  & 0.72  & 0.64 & 0.78  & \textbf{\textcolor{red}{0.83}} & \textbf{\textcolor{red}{0.80}} \\
B       & 0.58 & 0.56 & -- & 0.56  & 0.60  & -- & 0.73  & 0.69 & -- \\
C       & 0.56 & 0.55 & -- & 0.68  & 0.71  & -- & 0.58  & 0.55 & --  \\
D       & 0.74 & 0.77 & -- & 0.74  & 0.74 & -- & 0.74  & 0.77 & --  \\
E       & 0.76 & 0.77 & -- & 0.59  & 0.64  & -- & 0.78  & 0.83 & -- \\

\midrule 
\textbf{Average}  & 0.66 & 0.67 & 0.82 & 0.66  & 0.68  & 0.64 & 0.72  & 0.73 & 0.80 \\
\bottomrule
\end{tabular}
\vspace{5mm}
\caption{Performance of different permutations of the LLM-based phenotype pipeline in the validation set, measured in terms of $F_1$ score. We conducted iterations using five different prompt designs and employed three distinct aggregation methods in the MapReduce stage (see the Aggregation section in the paper). Furthermore, we conducted experiments involving the removal of echocardiogram (ECHO) and computerized tomography (CT) reports, both through regular expression mentions and by appending exclusion instructions to the LLM prompt. The results in red indicate the configurations selected for comparison with the structured phenotype in the test set.}
    \label{tab:validation_results}
\end{table}

\begin{table}[ht]
    \centering
    \begin{tabular}{ccccc}
        \toprule
        Prompt Design  & Steering & CoT & Multiple Choice \\
        \midrule

        \textbf{A} & \ding{51} & \ding{51}  & \ding{51}      \\
        \textbf{B} & \ding{51} & \ding{51}  &                \\
        \textbf{C} & \ding{51}* & \ding{51}    &             \\
        \textbf{D} & \ding{51} &            & \ding{51}      \\
        \textbf{E} & \ding{51} &            &                \\
        \midrule

    \end{tabular}
    \vspace{5mm}
    \caption{Descriptions of the various prompt templates explored during the evaluation. Steering column refers to steering the LLM to count history of PH as a "no" and possibly having PH as a "yes". (*) In template C, prompt was modified to remove the phrase "explain your reasoning"; which is included in all other prompt templates.}
    \label{tab:prompt_description}
\end{table}

\newpage

\subsection{Test Set Retrieval Note Type Distribution}
\label{sec:appendixA4}

\begin{table}[ht]
    \centering
    \begin{tabular}{l|c}
    \toprule
        Note Type & Frequency \\
            \midrule
        Progress Note &  $54.21\%$ \\
        Consult &  $7.95\%$ \\
        Discharge Summary &  $6.06\%$ \\
        History \& Physical Exam & $5.77\%$  \\
        Procedure &  $5.53\%$ \\
        Telephone Encounter &  $3.94\%$ \\
        Other &  $16.54\%$ \\
    \midrule
    \end{tabular}
    \vspace{5mm}
    \caption{Distribution of note types retrieved via Regex that are input into the LLM within the test set. “Other” consists of 23 different note types.}
    \label{tab:note_types}
\end{table}

\end{document}